\documentclass{article}
\PassOptionsToPackage{numbers,compress}{natbib}
\usepackage[preprint]{neurips_2026}
\usepackage[utf8]{inputenc}
\usepackage[T1]{fontenc}
\usepackage{hyperref}
\usepackage{url}
\usepackage{booktabs}
\usepackage{amsmath,amssymb,amsthm}
\usepackage{amsfonts}
\usepackage{graphicx}
\usepackage[protrusion=true,expansion=false]{microtype}
\usepackage{xcolor}
\usepackage{subcaption}
\usepackage{multirow}
\usepackage{array}
\usepackage{tikz}
\usepackage{float}
\graphicspath{{figures/}}
\newcommand{\Ncorpus}{35{,}309}
\newcommand{\Nsubj}{17{,}399}
\newcommand{\Ncohorts}{18}
\newcommand{\Nsites}{200{+}}
\title{BrainG3N: A Dual-Purpose Tokenizer for Controllable 3D Brain MRI Generation}
\author{%
  Max Van Puyvelde\,$^{*1,2}$\\
  \texttt{maxvpuyv@stanford.edu}\\
  \And
  H.~Ibrahim Gulluk\,$^{*3}$\\
  \texttt{gulluk@stanford.edu}\\
  \AND
  Wim Van Criekinge\,$^{\dagger 2}$\\
  \texttt{wim.vancriekinge@ugent.be}\\
  \And
  Olivier Gevaert\,$^{\dagger 1}$\\
  \texttt{ogevaert@stanford.edu}\\
  \AND
  \normalfont $^{1}$Department of Biomedical Data Science, Stanford University School of Medicine\\
  $^{2}$Department of Mathematical Modelling, Statistics \& Bioinformatics, Ghent University\\
  $^{3}$Department of Electrical Engineering, Stanford University\\
}
\begin{document}
\maketitle
\renewcommand{\thefootnote}{\fnsymbol{footnote}}
\footnotetext[1]{Joint first authors.}
\footnotetext[2]{Joint senior authors.}
\renewcommand{\thefootnote}{\arabic{footnote}}
\begin{abstract}
    Three-dimensional (3D) brain MRI is central to clinical neurology and neuro-oncology, where generative models could augment under-represented cohorts, simulate disease trajectories, and support privacy-preserving data sharing. Latent diffusion has been the go-to solution for modeling imaging data, but it places two competing demands on the tokenizer: encoder embeddings must retain the clinical information that downstream tasks act on, and the decoder must reconstruct anatomically faithful volumes. Existing reconstruction-driven tokenizers achieve the second at the expense of the first. To address this, we introduce a fully volumetric masked-autoencoder (MAE) based tokenizer for 3D brain MRI latent diffusion, decoupling encoder and decoder: a frozen 3D MAE encoder produces clinically informative embeddings, while a dedicated CNN decoder reconstructs voxels from a linear projection of those embeddings. We pretrain the encoder on \Ncorpus{} volumes from \Ncohorts{} public cohorts spanning four modalities, ten disease categories, and \Nsites{} acquisition sites, and demonstrate its dual utility in two settings. First, on a 23-task linear-probing benchmark, the encoder outperforms or matches SOTA models (i.e., BrainIAC, BrainSegFounder, and MedicalNet) on 21 of 23 tasks. Second, a conditional diffusion transformer (DiT) trained on these clinically informative embeddings supports both conditional generation across six variables and patient-specific longitudinal forecasting. Together these results establish a single 3D brain-MRI embedding space capable of both downstream clinical tasks and controllable generation.
    \end{abstract}
\begin{center}
  \small Model weights: \url{https://huggingface.co/gevaertlab/braing3n}
\end{center}
\section{Introduction}

In neurology and neuro-oncology, brain MRI informs clinical decisions ranging from tumor diagnosis and treatment planning to staging and monitoring of neurodegenerative diseases such as Alzheimer's and Parkinson's, and supports population-scale research on brain development and aging. Generative models on 3D brain MRI could extend this practice in several directions: augmenting under-represented patient cohorts, producing patient-specific digital twins~\cite{sadee2025digitaltwins} that simulate counterfactual disease trajectories, and enabling privacy-preserving cohort sharing across institutions where regulatory and logistical barriers currently prevent access to real imaging. Realizing these applications requires generative models at full 3D resolution, yet much of the field still operates on 2D slices. Because direct voxel-space generation is computationally infeasible at that scale, generative pipelines have broadly converged on latent diffusion~\cite{rombach2022ldm}: an encoder--decoder tokenizer first compresses volumes into a low-dimensional latent space, and a diffusion model is then trained on those embeddings. Conditional generation in this setting places two distinct demands on the tokenizer. First, the encoder embeddings must carry the clinical information needed for both conditional generation and downstream clinical tasks. Second, the decoder must reconstruct anatomically faithful voxel volumes. Existing 3D radiographic latent diffusion pipelines~\cite{pinaya2022brain,btb3d2025,meddifffm2024,med3ddiffusion2024,generatect2024} train a single encoder--decoder against a reconstruction objective, which biases the encoder toward voxel fidelity at the expense of clinical content; the resulting latent space is typically evaluated only on voxel-level reconstruction metrics.

We propose a dual-purpose self-supervised approach for 3D brain MRI in which a frozen masked-autoencoder (MAE) encoder produces an embedding space that serves two roles: a clinical representation for downstream tasks, and the feature space of a conditional diffusion transformer (DiT). To use this embedding space within a latent-diffusion pipeline, the encoder is paired with a CNN decoder via a linear projection of the embeddings (\S\ref{sec:method}). Using an MAE as a tokenizer for downstream diffusion has recent 2D precedent~\cite{chen2025maetok}; whether the same approach transfers to 3D radiographic data, where corpora are orders of magnitude smaller, the input dimensionality is higher, the relevant axes are subvisual phenotypes rather than visible object categories, is the question we address here.

Our work makes two main contributions. First, the frozen MAE encoder, pretrained on \Ncorpus{} brain MRI volumes from \Ncohorts{} cohorts, produces clinically informative embeddings: on a 23-task linear-probing benchmark (\S\ref{sec:exp-richness}) the frozen encoder outperforms or matches BrainIAC~\cite{brainiac2026}, BrainSegFounder~\cite{brainsegfounder2024}, and MedicalNet~\cite{chen2019medicalnet} on 21 of 23 tasks; for example, the encoder reaches AUC $0.937$ on isocitrate dehydrogenase 1 (IDH1) mutation status prediction, a key genomic biomarker for glioma diagnosis and treatment stratification; AUC $0.921$ on tumor grade classification; a mean absolute error of $4.43$ years on brain age regression; and AUC $0.967$ on sex prediction. Second, a conditional diffusion transformer (DiT)~\cite{peebles2023dit,lipman2023flow} trained on the same embeddings supports controllable generation across six variables (\S\ref{sec:exp-transfer}) and patient-specific longitudinal forecasting (\S\ref{sec:exp-longitudinal}); generated embeddings are mapped back to high-fidelity 3D voxel volumes by the CNN decoder. In both cases, generated samples are correctly recovered by classifiers trained on real data: for example, Pearson $r{=}0.93$ on cross-sectional age conditioning and Pearson $r{=}0.72$ on longitudinal age progression. This transfer test connects the embedding's encoding of clinical phenotypes to its generative controllability.

\section{Method}
\label{sec:method}

\paragraph{Pretraining corpus.} Our pretraining corpus comprises \Ncorpus{} brain MRI volumes from \Nsubj{} unique subjects across \Ncohorts{} public cohorts and \Nsites{} acquisition sites. Volumes span four modalities (T1, T2, fluid-attenuated inversion recovery [FLAIR], and T1 contrast-enhanced [T1c]) and ten clinical categories covering healthy controls, neurodegenerative disease (Alzheimer's, Parkinson's), neurodevelopmental conditions, psychiatric disorders, and brain tumors; subject ages range from 5 to 98 years. A single harmonized preprocessing pipeline registers every volume to the SRI24 atlas~\cite{rohlfing2010sri24} via ANTs affine registration~\cite{avants2008ants}, performs skull stripping with HD-BET~\cite{isensee2019hdbet}, and corrects intensity inhomogeneity with N4 bias-field correction~\cite{tustison2010n4}, producing $160{\times}192{\times}160$ volumes at $1$\,mm isotropic spacing. Cohort-specific entry points handle raw inputs at different preprocessing stages (already-stripped, defaced, native DICOM, etc.) without double-processing. The full dataset card and per-cohort pipelines are in Appendices~\ref{app:cohorts} and~\ref{app:preprocessing}.

\paragraph{MAE encoder.} The encoder is a 3D masked autoencoder~\cite{he2022mae} built on a 12-layer vision transformer~\cite{dosovitskiy2021vit} with hidden dimension 1152 and $16^{3}$ patches, producing 1200 tokens per volume. During pretraining, 70\% of patches are randomly masked: the encoder processes only the 360 visible patches, and a separate transformer decoder reconstructs the 840 masked patches from the encoder output under a per-patch mean-squared-error loss. Reconstructing 840 missing patches from 360 visible ones requires modeling long-range anatomical context, which forces the encoder to capture global structural relationships rather than local voxel statistics. This property is what we exploit downstream, both for clinical prediction and as the input space for the diffusion tokenizer.

\begin{figure}[!t]
  \centering
  \begin{subfigure}[b]{\linewidth}
    \raggedright\small (a) Two-phase tokenizer\par\vspace{0.1em}
    \includegraphics[width=\linewidth]{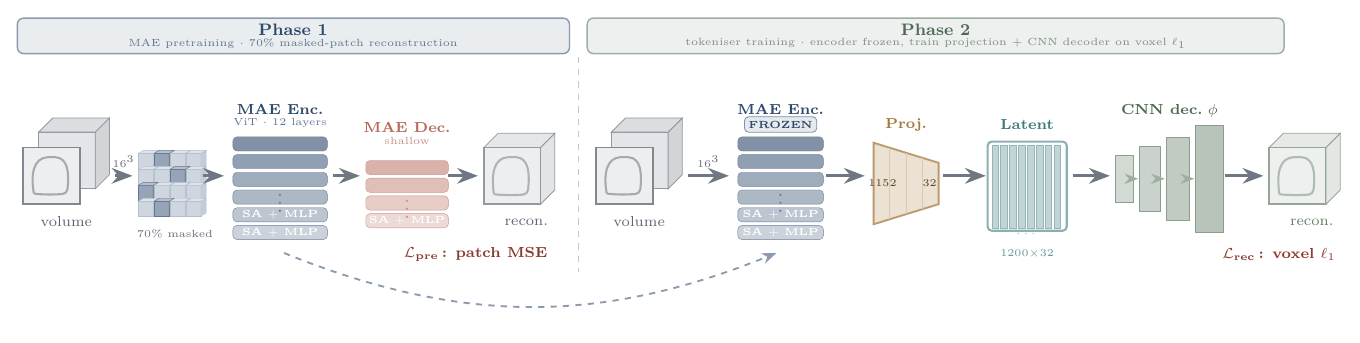}
    \phantomcaption
    \label{fig:arch-tok}
  \end{subfigure}
  \\[-1.4em]
  \begin{subfigure}[b]{\linewidth}
    \begin{tikzpicture}
      \node[anchor=south west, inner sep=0] (img) {\includegraphics[width=\linewidth]{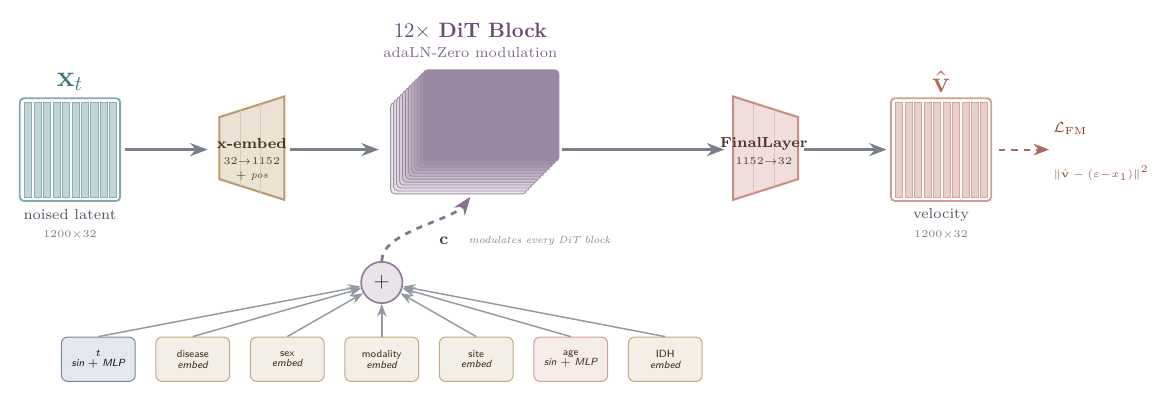}};
      \node[anchor=north west, font=\small] at ([xshift=2pt, yshift=-2pt]img.north west) {(b) Conditional flow-matching DiT};
    \end{tikzpicture}
    \phantomcaption
    \label{fig:arch-dit}
  \end{subfigure}
  \caption{Architecture. \textbf{(a)} Phase 1 pretrains a 3D MAE encoder on 70\% masked-patch reconstruction; Phase 2 freezes the encoder and trains a linear projection $P{\in}\mathbb{R}^{1152{\times}32}$ + 3D CNN decoder under voxel $\ell_{1}$. The same frozen feature space $z'{=}zP$ is consumed by the probe and produced by the DiT. \textbf{(b)} Noised tokens $\mathbf{x}_{t}$ pass through a 12-block DiT stack with adaLN-Zero modulation by a conditioning vector $\mathbf{c}$, producing the 32-channel velocity $\hat{\mathbf{v}}$. Categorical conditions use embedding lookups with a $K{+}1^{\text{st}}$ null slot for CFG dropout; age uses a sinusoidal $+$ MLP head.}
  \label{fig:arch}
\end{figure}

\paragraph{Two-phase tokenizer (Fig.~\ref{fig:arch}(a)).} The tokenizer couples the frozen MAE encoder to a 3D CNN decoder through a linear projection. The pretrained encoder is frozen; a linear projection $P \in \mathbb{R}^{1152 \times d'}$ compresses each of the 1200 tokens from $1152$ to $d'{=}32$ channels; and a 3D CNN decoder $\phi$ reconstructs voxels from the projected tokens under an $\ell_{1}$ loss:
\begin{equation}
  z = \mathrm{Enc}(x) \text{\;[frozen]}, \qquad z' = z P, \qquad \hat{x} = \phi(z').
\end{equation}

Jointly training encoder and decoder against the same reconstruction objective, as in CNN-VAE tokenizers, drifts the encoder toward local intensity fidelity and degrades its clinical content. The bottleneck $d'{=}32$ is a deliberate trade-off: a smaller bottleneck reduces the diffusion model's input dimensionality, while still preserving most of the encoder's clinical content (sweep in Table~\ref{tab:1100-validation}).

\paragraph{Linear probing.} We evaluate the clinical content of the frozen encoder embeddings via linear probing, the standard protocol in self-supervised representation learning: a single linear classifier is fit per task on top of the frozen embeddings, with no fine-tuning of the encoder. The input to each probe is the encoder's 1200-token output mean-pooled to a single $d{=}1152$ vector per volume. We use logistic regression for classification tasks and ridge regression for regression tasks, evaluated under 5-fold subject-grouped stratified cross-validation so that no subject appears in both the training and test splits of any fold. The full 23-task panel and per-modality breakdown are in Appendix~\ref{app:probe-full}.

\paragraph{Conditional latent diffusion (Fig.~\ref{fig:arch}(b)).} A flow-matching diffusion transformer (DiT)~\cite{peebles2023dit,lipman2023flow} is trained on the projected token sequences $z' \in \mathbb{R}^{1200 \times 32}$. The DiT has 12 blocks, hidden dimension $1152$, and $18$ attention heads, and is trained with the flow-matching objective
\begin{equation}
  \label{eq:dit}
  x_{t} = (1{-}t)\,x_{1} + t\,\varepsilon, \qquad \mathcal{L} = \lVert v_{\theta}(x_{t}, t, \mathbf{c}) - (\varepsilon - x_{1}) \rVert_{2}^{2},
\end{equation}
where $x_{1}$ is a real latent, $\varepsilon \sim \mathcal{N}(0, I)$, and the conditioning vector $\mathbf{c}$ is the sum of six condition embeddings routed through adaLN-Zero modulation~\cite{peebles2023dit}, $\mathrm{modulate}(h;\mathbf{c}) = h\,(1 + \mathrm{scale}(\mathbf{c})) + \mathrm{shift}(\mathbf{c})$. The six conditions are disease (8-way), sex, modality (4-way, never dropped), acquisition site (19-way), age (continuous), and IDH1 mutation status (binary). Classifier-free guidance (CFG)~\cite{ho2022cfg,dhariwal2021cfg} is implemented by independently replacing each condition with a null embedding at probability $p{=}0.1$ during training; at sampling time the velocity is extrapolated per condition as $v = v_{\text{uncond}} + s\,(v_{\text{cond}} - v_{\text{uncond}})$ with $s{=}2.0$ (CFG-scale sensitivity in Appendix~\ref{app:cfg-sweep}). Modality is never dropped because it is always specified at inference, so a null-modality branch would only waste capacity. The longitudinal variant of the DiT (\S\ref{sec:exp-longitudinal}) reuses the same frozen tokenizer and adaLN-Zero conditioning, replacing the noise-to-data interpolant with a baseline-to-followup latent bridge for patient-specific forecasting at requested time horizons.

\section{Experiments}
\label{sec:experiments}

Section~\ref{sec:exp-validation} validates the architectural choice on a small-scale benchmark; Sections~\ref{sec:exp-richness}--\ref{sec:exp-longitudinal} evaluate the resulting embeddings in three settings: cross-sectional probing, conditional generation, and longitudinal forecasting.

\subsection{Architectural validation at matched scale}
\label{sec:exp-validation}

Before scaling to the full corpus, we validated two architectural choices on a 1100-volume tumor cohort (UCSF-PDGM $+$ UPENN-GBM): the projection bottleneck $d'$, and the choice of MAE--CNN versus CNN-VAE tokenizer.

\paragraph{Bottleneck dimension $d'$.} We sweep the projection dimension over $d' \in \{32, 128, 512\}$ and train the Phase-2 decoder at each value. Reconstruction quality increases monotonically with $d'$, with diminishing returns above $d'{=}128$. To assess whether the projection preserves the clinical information that downstream tasks depend on, we additionally probe the projected features $z'$ at each $d'$ with linear classifiers. The IDH1 probing AUC at $d'{=}32$ is $0.861$, $0.022$ below the AUC obtained by probing the raw 1152-dim encoder embeddings ($0.883$) at less than 3\% of the total dimensionality (38K vs 1.4M); the corresponding gap on WHO tumor grade is $0.055$. We therefore adopt $d'{=}32$ for the remainder of the paper: the smaller bottleneck reduces the diffusion model's input dimensionality and training cost, while the linear projection preserves enough of the encoder's clinical content in $z'$ to support downstream conditioning (full sweep in Appendix~\ref{app:1k-baseline}, Table~\ref{tab:1100-validation}).

\begin{figure}[!t]
  \centering
  \includegraphics[width=0.95\linewidth]{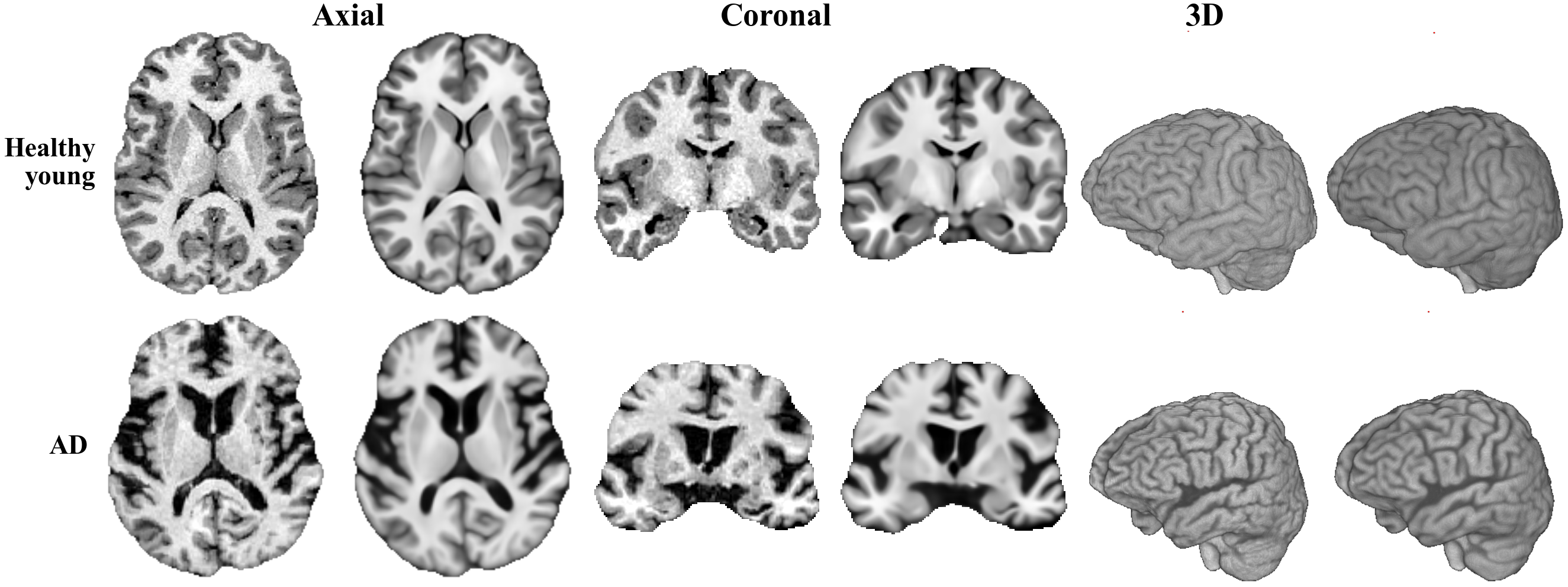}
  \caption{Voxel reconstructions from the frozen-MAE + CNN-decoder tokenizer at $d'{=}32$. Rows: a healthy young adult and an Alzheimer's case. Within each view (axial, coronal, 3D rendered), left = ground truth, right = reconstruction. Cortical folding, ventricular geometry, and overall morphology are preserved across age and disease state.}
  \label{fig:recon}
\end{figure}

\paragraph{Tokenizer architecture.} We compare the frozen MAE--CNN tokenizer against an AutoencoderKL (AKL)~\cite{pinaya2022brain}, the canonical 3D medical generative tokenizer, trained from scratch on the same cohort at matched compute. At the matched-dimensionality comparison ($d'{=}512$ vs AKL, both 614K total elements), MAE outperforms AKL on linear probing by $+0.064$ AUC on IDH1 and $+0.046$ AUC on WHO tumor grade. The same pattern obtains at smaller MAE bottlenecks: at our chosen $d'{=}32$ (38K vs AKL's 614K), MAE outperforms AKL on IDH1 ($+0.060$ AUC) and matches it on tumor grade. MAE is also more compute-efficient at training: the 70\% masking ratio means only $360$ of $1200$ tokens enter the encoder per step, supporting larger per-GPU batches than full-volume CNN-VAE training at the same memory budget. Based on this small-scale validation, we then pretrained the full MAE--CNN tokenizer on the entire \Ncorpus{}-volume corpus; this full-scale tokenizer is the one used throughout the rest of the paper, with representative reconstructions shown in Figure~\ref{fig:recon}.

\subsection{Frozen linear probing of clinical variables}
\label{sec:exp-richness}

\begin{table}[!t]
\centering
\small
\setlength{\tabcolsep}{4pt}
\caption{Head-to-head linear probing on the \Ncorpus{}-volume corpus. All four encoders evaluated on identical splits with identical probe code; the modality used per task (in parentheses) is fixed across encoders and chosen for clinical relevance. AUC ($\uparrow$) for classification, MAE ($\downarrow$) for regression, mean $\pm$ std across 5 subject-grouped folds. Shown: 8 above-floor tasks; full 23-task panel (best modality per encoder) in Appendix~\ref{app:competitors-full}. Abbreviations: CDR (Clinical Dementia Rating), MMSE (Mini-Mental State Examination).}
\label{tab:positioning}
\begin{tabular}{@{}llcrrrr@{}}
\toprule
Task (modality)              & Metric    & BrainIAC                & BSF                     & MedicalNet              & Ours                              \\
\midrule
IDH1 mutation (T1c)           & AUC       & 0.805\,$\pm$\,.042      & 0.780\,$\pm$\,.042      & 0.636\,$\pm$\,.042      & \textbf{0.937\,$\pm$\,.020}       \\
Tumor grade (T1)            & AUC       & 0.855\,$\pm$\,.029      & 0.799\,$\pm$\,.036      & 0.763\,$\pm$\,.051      & \textbf{0.921\,$\pm$\,.019}       \\
Disease 8-way (all)          & AUC       & 0.908\,$\pm$\,.004      & 0.891\,$\pm$\,.005      & 0.811\,$\pm$\,.006      & \textbf{0.948\,$\pm$\,.002}       \\
CDR staging (T1)             & AUC       & 0.626\,$\pm$\,.007      & 0.586\,$\pm$\,.008      & 0.599\,$\pm$\,.016      & \textbf{0.717\,$\pm$\,.025}       \\
Sex (T1)                     & AUC       & 0.787\,$\pm$\,.007      & 0.793\,$\pm$\,.005      & 0.613\,$\pm$\,.008      & \textbf{0.967\,$\pm$\,.004}       \\
Modality 4-way (all)         & AUC       & \textbf{1.000\,$\pm$\,.000} & \textbf{1.000\,$\pm$\,.000} & 0.991\,$\pm$\,.001 & \textbf{1.000\,$\pm$\,.000}       \\
\midrule
Brain age (T1)               & MAE (y)   & 7.33\,$\pm$\,.11        & 8.90\,$\pm$\,.15        & 12.54\,$\pm$\,.17       & \textbf{4.43\,$\pm$\,.07}         \\
MMSE (all)                   & MAE       & 2.16\,$\pm$\,.03        & 2.36\,$\pm$\,.06        & 2.29\,$\pm$\,.06        & \textbf{1.91\,$\pm$\,.03}         \\
\bottomrule
\end{tabular}
\end{table}

\paragraph{Head-to-head vs published brain-MRI foundation models.} We re-ran frozen-feature linear probes for the three published 3D brain-MRI foundation models with public encoders, BrainIAC~\cite{brainiac2026}, BrainSegFounder~\cite{brainsegfounder2024}, and MedicalNet~\cite{chen2019medicalnet}, on our \Ncorpus{}-volume corpus using identical splits and probe code; a more recent fourth, BrainDINO~\cite{wu2026braindino}, has not released encoder weights or pretraining code. On 8 representative clinical and acquisition-related tasks (Table~\ref{tab:positioning}), our frozen encoder outperforms all three competitors on 7 of 8 tasks; on the 8th (modality recovery) all encoders are essentially at ceiling. Across the full 23-task panel (Appendix~\ref{app:competitors-full}, 15 classification + 8 regression), we outperform or match competitors on 21 of 23 tasks; the two non-wins, Geriatric Depression Scale (GDS) and the third part of the Unified Parkinson's Disease Rating Scale (UPDRS-III), are near-floor regression tasks for every encoder. As a sanity check, our reproduction of BrainIAC's frozen-feature brain-age MAE on T1 is $7.33$\,y, close to its published $7.51$\,y, confirming the implementation follows BrainIAC's protocol. The most recent brain-MRI foundation model, BrainDINO, is a 2D slice-based DINOv3-style model; the authors report brain-age MAE $5.54$\,y and IDH1 AUC $0.901$ at 100\% supervision under a lightweight task-head fine-tune, vs $4.43$\,y and $0.937$ from our frozen linear probes on the corresponding tasks, positioning it below our fully volumetric self-supervised approach.

\subsection{Probe-to-controllability transfer}
\label{sec:exp-transfer}

\begin{table}[t]
\centering
\caption{DiT generation evaluation under CFG ($s{=}2.0$). \textbf{(a)} Controllability: real-data probes applied to generated samples. \emph{Real}: probe AUC on real held-out data (Pearson $r$ for age), the ceiling. \emph{Cond}: agreement with the requested class. \emph{Null}: probe output on CFG-null samples. \emph{Prior}: dataset class frequency. Modality is never dropped during training, hence no null branch. Disease, sex, and age use T1; IDH1 uses T1c (glioblastoma [GBM] subset). \textbf{(b)} 3D-FID per condition and pooled across 14 arms; per-arm decomposition in Appendix~\ref{app:gen}. Disease classes: HC = healthy control, AD = Alzheimer's disease, MCI = mild cognitive impairment, PD = Parkinson's disease.}
\label{tab:gen}

\begin{subtable}[t]{0.66\linewidth}
\raggedright
\caption{Cross-sectional controllability}
\label{tab:control}
\setlength{\tabcolsep}{4pt}
\small
\begin{tabular}{@{}llcccc@{}}
\toprule
Condition & Task                  & Real          & Cond         & Null              & Prior         \\
\midrule
Disease   & HC vs AD              & 0.99          & 0.99         & 0.92 HC           & 0.91 HC       \\
Disease   & HC vs MCI             & 0.90          & 0.80         & 0.67 MCI          & 0.70 HC       \\
Disease   & MCI vs AD             & 0.76          & 0.90         & 0.97 MCI          & 0.80 MCI      \\
Disease   & HC vs PD              & 0.96          & 0.82         & 0.97 HC           & 0.87 HC       \\
Sex       & F vs M                & 0.97          & 0.93         & 0.72 F            & 0.43 F        \\
Modality  & 4-way                 & 1.00          & 1.00         & ---               & 0.25          \\
Age       & 30\,/\,50\,/\,70\,y   & $r{=}0.98$    & $r{=}0.93$   & $\hat{a}{=}55.5$  & mean $55.5$   \\
IDH1       & wt vs mut             & 0.94          & 0.52         & 1.00 wt           & 0.88 wt       \\
\bottomrule
\end{tabular}
\end{subtable}%
\hfill
\begin{subtable}[t]{0.30\linewidth}
\raggedleft
\caption{3D-FID ($\downarrow$)}
\label{tab:fid-pooled}
\setlength{\tabcolsep}{4pt}
\small
\begin{tabular}{@{}lcc@{}}
\toprule
Condition & vs recons & vs real \\
\midrule
Disease   & 28.6      & 136.3   \\
Sex       & 46.0      & 116.3   \\
Modality  & 47.3      & 116.0   \\
Age       & 29.8      & 122.6   \\
IDH1       & 19.0      & 132.5   \\
\midrule
Pooled    & 34.4      & 107.3   \\
\bottomrule
\end{tabular}
\end{subtable}

\end{table}

\paragraph{Protocol.} For each DiT condition $c$, we (i) train a probe for $c$ on real volumes, (ii) draw 64 samples from the DiT under each requested class of $c$ and a matching set under null conditioning, (iii) apply the frozen real-data probe to both sets. A controllable condition recovers the requested attribute under \textsc{cond} and falls back to the class prior (or dataset mean) under \textsc{null}; the \textsc{cond}--\textsc{null} gap measures the conditional signal the sampler preserves.

\begin{figure}[!t]
  \centering
  \includegraphics[width=\linewidth]{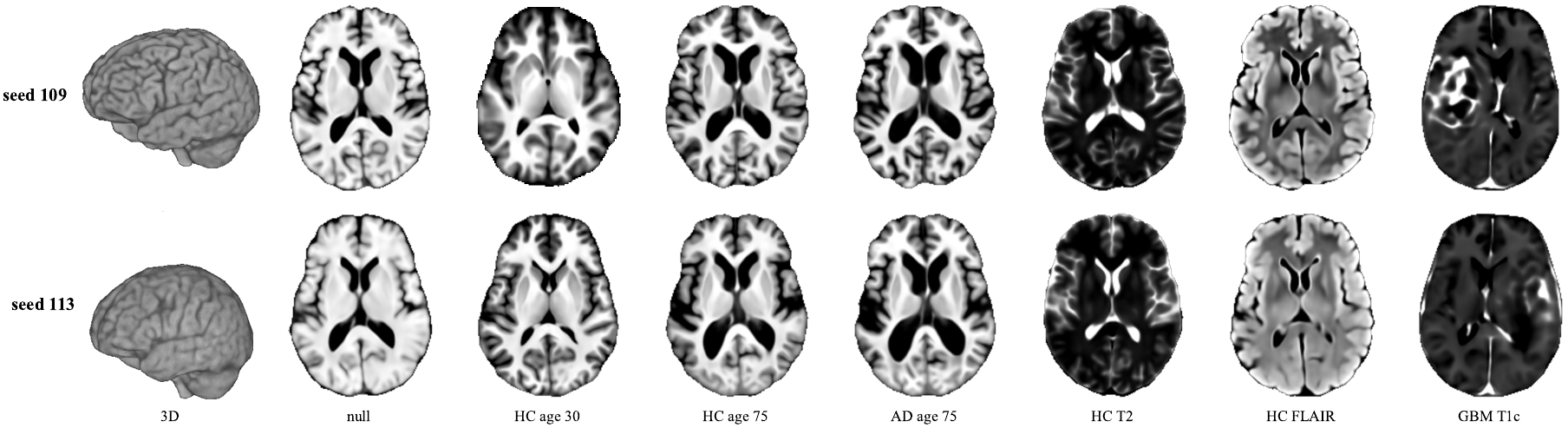}
  \caption{Same-noise counterfactual generation under the conditional DiT (CFG $s{=}1.5$). Each row fixes the initial latent noise (seeds 109 and 113); columns sweep one set of conditioning attributes at a time. Leftmost column: 3D rendering of the unconditioned baseline. The remaining seven columns sweep \emph{age} (HC, 30 vs.\ 75\,y), \emph{disease} (HC vs.\ AD at 75\,y), \emph{modality} (T1, T2, FLAIR), and \emph{tumor} (GBM IDH1-wildtype on T1c). Within a row, identity-preserving anatomy is largely retained while the swapped attribute drives the visible change.}
  \label{fig:counterfactual}
\end{figure}

\paragraph{Results.} Real-data probes reliably recover the requested clinical attributes from samples generated under classifier-free guidance (Table~\ref{tab:control}). Disease HC-vs-AD recovers at 0.99 agreement under \textsc{cond} and collapses to the class prior under \textsc{null} (92\% HC vs 91\% prior); control extends to the other three disease contrasts (MCI vs AD: Cond 0.90; HC vs MCI: 0.80; HC vs PD: 0.82). Under \textsc{null}, HC vs PD and MCI vs AD collapse to the majority class as expected, while HC vs MCI shows a residual MCI bias (0.67 MCI vs 0.30 MCI prior), suggesting the unconditional model preferentially produces disease-leaning samples on this contrast. Sex reaches 0.93 against a 0.72 \textsc{null} baseline; modality, which is never dropped during training, is perfectly recovered (cond agreement $1.00$ across the four classes). Age tracks the requested sweep at Pearson $r{=}0.93$, with predicted means $35.4\,/\,54.4\,/\,76.0$ for requested $30\,/\,50\,/\,70$ years and null generations centered at the dataset mean of $55.5$\,y. IDH1 is the hardest axis (mean cond agreement $0.52$ against an 88\% wildtype prior; balanced-probe variant $0.56$), reflecting that classifier-free guidance struggles to steer toward the rare mutant class on the small T1c tumor subset rather than a probe artefact; the null branch collapses fully to wildtype. Figure~\ref{fig:counterfactual} shows same-noise counterfactual sweeps: holding the noise seed fixed and varying one conditioning attribute at a time produces the expected visible change (ventricular enlargement under AD, modality contrast switching, GBM enhancing tumor mass on T1c) while preserving the overall anatomical layout.

\paragraph{Generation fidelity.} Pooled across all 14 conditional arms, the FID between generated samples and the tokenizer's reconstructions (\emph{gen-vs-recons}) is $34.4$, and between generated samples and raw real volumes (\emph{gen-vs-real}) is $107.3$ (Table~\ref{tab:control}; $n_{\text{real}}{=}2000$ vs $n_{\text{gen}}{=}1088$). Both are competitive with prior 3D medical generative work (FIDs $40$--$120$~\cite{pinaya2022brain}). The tokenizer's own reconstruction floor against the real volumes (\emph{recons-vs-real}) is $77.4$; since gen-vs-recons is tighter than this floor, most of the gen-vs-real gap reflects the tokenizer's compression of high-frequency scanner-specific detail rather than generator quality. A sweep over the CFG scale $s$ (Appendix~\ref{app:cfg-sweep}) confirms $s{=}2.0$ sits on a near-optimal plateau, and a nearest-neighbor audit in latent space~\citep{carlini2023extracting} finds no evidence of training-set memorization: generated latents sit $1.55\times$ farther from their nearest training latent than training latents do from each other, with zero of $1088$ generated samples falling below the $5^{\text{th}}$-percentile real-to-real neighbor distance (Appendix~\ref{app:gen}).

\subsection{Patient-specific longitudinal forecasting}
\label{sec:exp-longitudinal}

Sections \ref{sec:exp-richness} and \ref{sec:exp-transfer} evaluated the frozen embeddings cross-sectionally and as the input space for conditional generation. We now test whether the same frozen tokenizer supports intra-patient temporal extrapolation: given a single baseline latent and a requested $\Delta t$, does the model produce a follow-up scan whose predicted brain age, as read by a real-data brain-age probe, increases by $\Delta t$?

\paragraph{Setup.} Using the frozen tokenizer of \S\ref{sec:method} unchanged, we train a longitudinal flow-matching DiT on $\sim25$k ADNI T1 baseline--followup latent pairs (subject-level split, 2.9k validation pairs). The model is conditioned on diagnosis (healthy control, mild cognitive impairment, or Alzheimer's disease), sex, baseline age, Clinical Dementia Rating (CDR) score, and the time horizon $\Delta t$ in years. The training interpolant couples baseline $x^{b}$ and follow-up $x^{t_f}$ with a vanishing-endpoint Brownian envelope~\cite{albergo2023stochastic}:
\begin{equation*}
  x_{t} \;=\; \underbrace{(1{-}t)\,x^{b} + t\,x^{t_f}}_{\text{interpolant}} \;+\; \underbrace{\sigma\sqrt{t(1{-}t)}\,\varepsilon}_{\text{envelope},\;\sigma=0.5}.
\end{equation*}
The interpolant alone admits a closed-form shortcut that lets the velocity ignore $\Delta t$; the envelope is zero at both endpoints (so $x_{t}$ hits $x^{b}$ at $t{=}1$ and $x^{t_f}$ at $t{=}0$ in expectation) but non-zero in the interior, breaking the shortcut and forcing the velocity to use $\Delta t$. Sampling is deterministic Euler integration; the noise is training-only. We use CFG $s{=}1.0$ rather than $s{=}2.0$ as in \S\ref{sec:exp-transfer}, since higher guidance scales amplify bridge-sampler artefacts in CSF; $s{=}1.0$ was selected based on visual quality (CFG sweep in Appendix~\ref{app:cfg-sweep}).

\begin{figure}[t]
  \centering
  \includegraphics[width=0.78\linewidth]{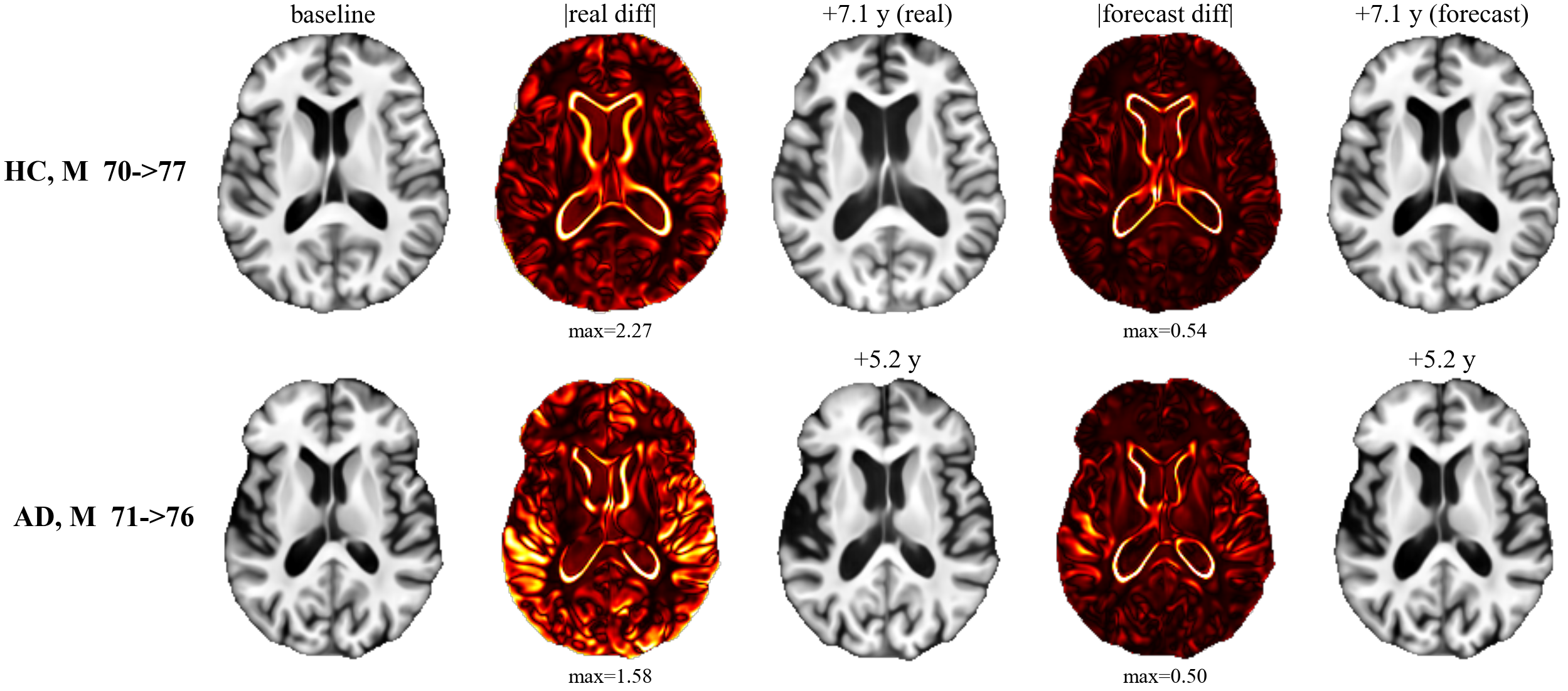}
  \caption{Real vs model-forecast longitudinal change for held-out ADNI subjects. Each row: baseline; $|$real diff$|$; real follow-up at $\Delta t$; $|$forecast diff$|$; model forecast at $\Delta t$. All volumes pass through the $d'{=}32$ MAE--CNN tokenizer. Diff cells use independent intensity scaling (hot colormap); absolute magnitudes (z-score) printed beneath. Forecast diffs capture the same anatomical loci (ventricle, sulci edges) as real diffs at damped magnitude (slope $0.268$ in §\ref{sec:exp-longitudinal}).}
  \label{fig:realvsforecast}
\end{figure}

\paragraph{Evaluation and results.} For each of 64 held-out validation baselines (no subject overlap with training), we forecast latents at $\Delta t \in \{0,1,2,5,10\}$\,y under classifier-free guidance ($s{=}1.0$, 100-step Euler integration), decode through the frozen tokenizer, and apply a brain-age probe trained on ADNI T1 (mean absolute error of $3.97$\,years on held-out cross-sectional data). The model recovers approximately 27\% of true aging in magnitude (slope $0.268$, Pearson $r{=}0.716$ between predicted and requested $\Delta t$), with the anatomical change correctly localized to ventricles and sulci (Figure~\ref{fig:realvsforecast}). The magnitude attenuation is consistent with the stochastic-bridge regularizer interpolating toward the conditional mean. Per-$\Delta t$ sweep visualizations are in Appendix~\ref{app:longit-sweep}.

\paragraph{Scope.} Specialised longitudinal brain-MRI LDMs~\cite{puglisi2024brlp,agldm2026} target volumetric ROI fidelity under task-specific losses and use CNN-VAE tokenizers. In contrast, the longitudinal variant here reuses the same frozen self-supervised embeddings used for cross-sectional probing and conditional generation, showing that this single embedding space also supports intra-patient temporal forecasting without retraining the encoder.

\section{Discussion}
\label{sec:discussion}

Our work makes two main contributions. First, a self-supervised two-phase MAE-CNN tokenizer for 3D brain MRI in which a frozen MAE encoder is decoupled from a dedicated CNN decoder: the encoder produces clinically informative embeddings (outperforming or matching the three published 3D brain-MRI foundation baselines on 21 of 23 linear-probing tasks; the two non-wins, GDS depression and UPDRS-III, are near-floor regression for every encoder), and the decoder reconstructs voxels faithfully. Second, a conditional flow-matching DiT trained on those frozen embeddings supports controllable generation: real-data probes applied to DiT samples recover the requested condition under classifier-free guidance, with \textsc{cond}--\textsc{null} gaps consistent with each probe's real-data signal. This supports our hypothesis that the MAE serves as a viable tokenizer for downstream diffusion in 3D brain MRI.

Both contributions are delivered by a single MAE encoder, MAE-CNN tokenizer, and DiT that span all four structural MRI modalities (T1, T2, FLAIR, T1c) across pretraining, probing, and conditional generation, in contrast to alternative approaches that rely on 2D slicewise pretraining or train separate per-modality models.

\paragraph{Broader impact.} The variables recovered by the linear probes (\S\ref{sec:exp-richness}) drive routine clinical decisions in neurology and neuro-oncology that today rely on invasive or costly ancillary procedures: for instance, IDH1 mutation status currently requires tumor biopsy, apolipoprotein E (APOE) genotyping requires blood draw, and cognitive scores such as CDR, MMSE, and the Montreal Cognitive Assessment (MoCA) require dedicated examiner time. Predicting these directly from MRI could therefore provide an image-only readout, with two operating points: linear probing on frozen embeddings (no GPU required at inference; sites without dedicated infrastructure can fit and apply task-specific classifiers on cached features) or supervised fine-tuning for higher performance when GPU compute is available. Brain-age regression similarly yields a continuous biomarker (the brain-age gap) that could serve as an indicator of accelerated neurodegeneration risk. The same variables condition the DiT (\S\ref{sec:exp-transfer}), allowing patient-specific longitudinal forecasting (\S\ref{sec:exp-longitudinal}), counterfactual generation by changing conditioning labels (e.g., ``what would this patient look like under alternative disease status''), and clinically grounded synthetic data for augmentation of under-represented patient cohorts and privacy-preserving cohort sharing.

\paragraph{Limitations.} Findings are established on brain MRI; transfer to other 3D modalities (CT, whole-body MRI, ultrasound) is untested. On longitudinal forecasting (\S\ref{sec:exp-longitudinal}), the bridge sampler recovers the trajectory direction and anatomical loci of true aging but only $\sim$27\% of its magnitude (slope $0.268$), consistent with the stochastic-bridge regularizer interpolating toward the conditional mean. The AKL comparison that motivates the two-phase choice is at 1100 volumes (Appendix~\ref{app:1k-baseline}); the architectural verdict at \Ncorpus{}-volume scale rests on the head-to-head against published foundation models rather than a re-run AKL. Pretraining distributions differ across the published baselines, with cohort overlap with our corpus ranging from substantial to none, so a portion of the head-to-head gaps conflates architecture and method with distribution mismatch. Acquisition confounders are detectable on the embeddings (field-strength and scanner-vendor probes exceed 0.94 AUC), so cohort-matched evaluation is required for any downstream application.

\paragraph{Reproducibility.}
Model weights (encoder, CNN decoder, and conditional DiT), together with loading and
feature-extraction code, are available at \url{https://huggingface.co/gevaertlab/braing3n}
under a non-commercial research-use license. Approximate total training: 200 H100-hours; probing runs on a 16-core CPU in $\approx 2$\,h per task. Hyperparameters in Appendix~\ref{app:hparams}.

\bibliographystyle{plainnat}
\bibliography{references}
\clearpage
\appendix
\section{Dataset card}
\label{app:cohorts}

\paragraph{Aggregation.} \Ncorpus{} preprocessed brain MRI volumes from \Nsubj{} unique subjects across \Ncohorts{} public cohorts and \Nsites{} acquisition sites worldwide. Four imaging modalities (T1, T2, FLAIR, T1c). Ages 5--98. 6{,}576 subjects have longitudinal (multi-visit) scans.

\begin{table}[H]
\centering
\scriptsize
\caption{Per-cohort composition. Dominant diagnoses listed; full per-subject labels in \texttt{volumes.csv}. Diagnosis abbreviations: HC (healthy control), AD (Alzheimer's disease), MCI (mild cognitive impairment), PD (Parkinson's disease), GBM (glioblastoma), ASD (autism spectrum disorder), ADHD (attention-deficit/hyperactivity disorder), SCZ (schizophrenia).}
\label{tab:cohorts}
\begin{tabular}{lrrll}
\toprule
Cohort            & Volumes  & Subjects & Modalities          & Dominant diagnoses  \\
\midrule
ADNI~\cite{petersen2010adni}          & 10{,}910 & 2{,}592  & T1                  & HC, MCI, AD         \\
PPMI~\cite{marek2011ppmi}             &  3{,}908 & 1{,}484  & T1, T2, FLAIR       & HC, PD              \\
NKI-RS~\cite{nooner2012nki}           &  2{,}455 & 1{,}326  & T1                  & HC                  \\
UPENN-GBM~\cite{bakas2022upenngbm}    &  2{,}444 &   611    & T1/T2/FLAIR/T1c     & GBM                 \\
HCP-YA~\cite{vanessen2013hcp}         &  2{,}226 & 1{,}113  & T1, T2              & HC                  \\
HBN~\cite{alexander2017hbn}           &  2{,}135 & 2{,}135  & T1                  & Paediatric          \\
UCSF-PDGM~\cite{calabrese2022ucsfpdgm}&  1{,}980 &   495    & T1/T2/FLAIR/T1c     & GBM, Glioma         \\
BGSP~\cite{holmes2015bgsp}            &  1{,}636 & 1{,}568  & T1                  & HC                  \\
ABIDE-II~\cite{dimartino2017abideII}  &  1{,}430 & 1{,}082  & T1                  & HC, ASD             \\
FCON1000~\cite{biswal2010fcon1000}    &  1{,}197 & 1{,}197  & T1                  & HC                  \\
IXI~\cite{ixi}                        &  1{,}159 &   582    & T1, T2              & HC                  \\
ABIDE-I~\cite{di2014abide}            &    984   &   947    & T1                  & HC, ASD             \\
SCHIZO~\cite{aine2017cobre,gollub2013mcic,wang2016schizconnect}
                                      &    670   &   335    & T1, T2              & HC, SCZ             \\
ADHD-200~\cite{adhd200consortium2012} &    598   &   598    & T1                  & HC, ADHD            \\
CoRR~\cite{zuo2014corr}               &    546   &   546    & T1                  & HC                  \\
OASIS-1~\cite{marcus2007oasis}        &    436   &   416    & T1                  & HC, MCI, AD         \\
OASIS-2~\cite{marcus2010oasis2}       &    373   &   150    & T1                  & HC, MCI, AD         \\
NKI-RS 2~\cite{nooner2012nki}         &    222   &   222    & T1                  & HC                  \\
\midrule
Total             & \textbf{\Ncorpus{}} & \textbf{\Nsubj{}} & 4 modalities & 10 classes \\
\bottomrule
\end{tabular}
\end{table}

\paragraph{Disease coverage.} HC 15{,}274, MCI 5{,}808, GBM 4{,}028, PD 3{,}381, paediatric-mixed 2{,}130, AD 1{,}458, ASD 1{,}061, Glioma (non-GBM) 396, SCZ 328, ADHD 247. Healthy subjects dominate (43\% of the corpus) by design: the probing benchmark requires a large HC pool to separate clinical signal from healthy-brain variability, and the generative model benefits from a well-represented null distribution.

\paragraph{Clinical metadata.} Coverage of key fields: age-at-scan 90\%, sex 90\%, dx 96\%, CDR 34\% (99\% within AD/MCI), MMSE 41\% (99\% within AD/MCI), MoCA 18\%, APOE 41\%, tumor grade 8\% (100\% within tumor cohorts), IDH1 7\% (91\% within tumor cohorts), MGMT 6\% (69\% within tumor cohorts), site 78\%.

\section{Preprocessing pipeline}
\label{app:preprocessing}

All volumes pass through a single harmonized pipeline before entering the model. Raw NIfTI/DICOM inputs first undergo N4 bias-field correction~\cite{tustison2010n4} to remove low-frequency intensity inhomogeneity from RF coil sensitivity profiles. HD-BET~\cite{isensee2019hdbet} then performs skull stripping, removing skull, meninges, and extracranial tissue. The resulting skull-stripped brain is affine-registered to the SRI24 atlas~\cite{rohlfing2010sri24,avants2008ants} at $240{\times}240{\times}155$ voxels, $1$\,mm isotropic, LPS orientation. For multi-modality studies, T1 is registered as the ``center'' modality and T2/FLAIR/T1c are co-registered through T1's transform (avoiding independent re-registration that would create cross-modal alignment drift). Negative voxels from registration interpolation are clipped to zero. Intensity normalization is not applied at preprocessing time; per-volume z-scoring is performed in the training dataloader. For modeling, volumes are center-cropped/padded to $160{\times}192{\times}160$.

\section{Hyperparameters}
\label{app:hparams}

\begin{table}[H]
\centering
\scriptsize
\caption{Training hyperparameters for each stage.}
\label{tab:hparams}
\resizebox{\linewidth}{!}{%
\begin{tabular}{p{3.6cm}p{3.0cm}p{6.8cm}}
\toprule
Component                    & Hyperparameter            & Value                                                     \\
\midrule
\multirow{8}{3.6cm}{\raggedright MAE encoder (Phase 1)}
    & Architecture              & ViT, 12 layers, hidden $1152$, MLP $4608$, 16 heads      \\
    & Decoder                   & ViT, 8 layers, hidden $1152$, MLP $4608$, 16 heads       \\
    & Patch size                & $16^{3}$; 1200 tokens / volume                           \\
    & Masking ratio             & 0.70                                                     \\
    & Batch / GPUs              & $16$ / 4$\times$H100                                     \\
    & Optimizer                 & AdamW, lr $1{\times}10^{-4}$, cosine                     \\
    & Epochs                    & 100                                                      \\
    & Final train / val loss    & $0.117$ / $0.121$ patch MSE                              \\
\midrule
\multirow{6}{3.6cm}{\raggedright Phase 2 tokenizer (frozen enc.\ $+$ proj.\ $+$ CNN dec.)}
    & Projection                & linear $P{\in}\mathbb{R}^{1152{\times}32}$               \\
    & CNN decoder               & ResBlocks $+$ attn, $4{\times}{\uparrow}2$, 43.7M params \\
    & Loss                      & voxel $\ell_{1}$                                         \\
    & Batch / GPUs              & $12$ / 4$\times$H100, bf16                               \\
    & Optimizer                 & AdamW, lr $1{\times}10^{-4}$                             \\
    & Epochs                    & 80, early-stop patience 40                               \\
\midrule
\multirow{7}{3.6cm}{\raggedright Conditional DiT (\S\ref{sec:exp-transfer})}
    & Architecture              & DiT-L: 12 blocks, hidden $1152$, $18$ heads, MLP $4{\times}$ \\
    & Conditions                & 6 (disease, sex, modality [never-drop], site, age, IDH1) \\
    & CFG dropout               & per-condition $p=0.1$                                    \\
    & Batch / GPUs              & $28$ / 4$\times$H100, bf16                               \\
    & Optimizer                 & AdamW, lr $5{\times}10^{-5}$, wd $0.01$, clip $0.5$      \\
    & EMA decay                 & $0.9999$                                                 \\
    & Sampling                  & Euler ODE, 50 steps, CFG $s{=}2.0$                       \\
\midrule
\multirow{8}{3.6cm}{\raggedright Longitudinal DiT (\S\ref{sec:exp-longitudinal})}
    & Architecture              & shares Conditional-DiT backbone (12 blocks, hidden $1152$, $18$ heads) \\
    & Training data             & $\sim25$k ADNI T1 baseline--followup latent pairs (subject-level split) \\
    & Conditions                & 5 (disease, sex, baseline age, CDR, $\Delta t$ in years) \\
    & Interpolant               & vanishing-endpoint Brownian bridge, $\sigma{=}0.5$~\cite{albergo2023stochastic} \\
    & CFG dropout               & per-condition $p=0.1$ ($\Delta t$ and disease never-drop) \\
    & Batch / GPUs              & $28$ / $1\times$H100, bf16                               \\
    & Optimizer                 & AdamW, lr $5{\times}10^{-5}$, wd $0.01$, clip $0.5$, EMA $0.9999$ \\
    & Sampling                  & deterministic Euler ODE, 100 steps, CFG $s{=}1.0$, EMA epoch-75 \\
\midrule
\multirow{5}{3.6cm}{\raggedright Probing}
    & Features                  & frozen encoder, mean-pool over 1200 tokens, $d{=}1152$  \\
    & Classifier                & logistic regression (\texttt{max\_iter=1000})            \\
    & Regressor                 & ridge ($\alpha=1.0$)                                     \\
    & CV                        & 5-fold stratified group $k$-fold (subject-grouped)       \\
    & Seed                      & 2025                                                     \\
\bottomrule
\end{tabular}%
}
\end{table}

Approximate total training cost: $\approx 200$ H100-hours (100 epochs MAE pretraining $+$ 80 epochs Phase 2 $+$ $\approx 1100$ DiT epochs). Probing runs on a 16-core CPU in $\approx 2$\,h per task.

\section{AKL baseline training details}
\label{app:1k-baseline}

The AutoencoderKL (AKL) baseline reported in Table~\ref{tab:1100-validation} is the canonical CNN-VAE tokenizer architecture from the 3D medical generative literature~\cite{pinaya2022brain}. We train AKL from scratch on the same 1100-volume tumor cohort (UCSF-PDGM $+$ UPENN-GBM) as the MAE+CNN tokenizer, with reconstruction $+$ KL loss; the resulting latent has 614K total elements, matching the MAE+CNN $d'{=}512$ configuration. Held-out reconstruction metrics are reported on the 20\% test split; clinical probing uses 5-fold subject-grouped cross-validation with logistic regression.

\begin{table}[H]
\centering
\small
\caption{Architectural validation on the 1100-volume tumor cohort: MAE+CNN at projection bottlenecks $d'$ vs an AutoencoderKL (AKL) baseline (matched compute). Reconstruction on held-out volumes; probing AUC under 5-fold logistic regression. Bold: best probing per column.}
\label{tab:1100-validation}
\begin{tabular}{lrrrrrr}
\toprule
                            & Dims  & MSE $\downarrow$ & PSNR $\uparrow$ & SSIM $\uparrow$ & IDH1 AUC $\uparrow$ & Grade AUC $\uparrow$ \\
\midrule
AKL ($d{=}8$)               & 614K  & 0.093            & 33.62           & 0.868           & 0.801              & 0.793                \\
\midrule
MAE ($d'{=}32$, chosen)     & 38K   & 0.119            & 32.54           & 0.859           & 0.861              & 0.791                \\
MAE ($d'{=}128$)            & 154K  & 0.075            & 34.60           & 0.912           & 0.877              & 0.827                \\
MAE ($d'{=}512$)            & 614K  & 0.066            & 35.16           & 0.921           & 0.865              & 0.839                \\
MAE ($d'{=}1152$, no proj.) & 1.4M  & 0.057            & 35.79           & 0.932           & \textbf{0.883}     & \textbf{0.846}       \\
\bottomrule
\end{tabular}
\end{table}

\paragraph{Reconstruction-vs-probing trade-off.} On voxel reconstruction, AKL is competitive with MAE+CNN at $d'{=}32$ (PSNR 33.62 vs 32.54), but MAE exceeds AKL at $d'{\geq}128$ (Table~\ref{tab:1100-validation}). On clinical probing, MAE outperforms AKL on IDH1 at every bottleneck and on WHO tumor grade at every bottleneck above $d'{=}32$, where the two are essentially tied. The MAE-as-tokenizer advantage is driven by what the encoder embedding captures rather than by raw voxel detail preservation.

\section{Full 23-task probing breakdown (per modality)}
\label{app:probe-full}

\begin{table}[H]
\centering
\scriptsize
\caption{Frozen linear-probe performance per task and modality split, \Ncorpus{}-volume corpus, 5-fold stratified group $k$-fold grouped by subject. AUC for classification, $R^{2}$ for regression. Dashes: too few samples in that modality subset. Bold $=$ best modality per task. Abbreviations: APOE (apolipoprotein E), IDH1 (isocitrate dehydrogenase 1), MGMT (O$^{6}$-methylguanine-DNA methyltransferase promoter methylation), CDR (Clinical Dementia Rating), MMSE (Mini-Mental State Examination), MoCA (Montreal Cognitive Assessment), GDS (Geriatric Depression Scale), UPDRS-III (Unified Parkinson's Disease Rating Scale, motor), SCOPA-AUT (Scale for Outcomes in PD-Autonomic).}
\label{tab:probe-full}
\resizebox{\linewidth}{!}{%
\begin{tabular}{lcccccc}
\toprule
Task                                & all                 & T1                  & T1c                 & T2                  & FLAIR               & patient-pooled      \\
\midrule
Disease (8-way, AUC)                & 0.948               & 0.919               & 0.959               & \textbf{0.997}      & 0.967               & 0.972               \\
Sex (AUC)                           & 0.961               & \textbf{0.967}      & 0.924               & 0.946               & 0.957               & 0.944               \\
Modality (4-way, AUC)               & \textbf{1.000}      & ---                 & ---                 & ---                 & ---                 & 1.000               \\
Site (19-way, AUC)                  & 0.998               & 0.996               & \textbf{1.000}      & \textbf{1.000}      & \textbf{1.000}      & 0.999               \\
Field strength (binary, AUC)        & \textbf{0.986}      & 0.986               & ---                 & ---                 & ---                 & 0.980               \\
Handedness (binary, AUC)            & 0.595               & 0.596               & ---                 & 0.500               & 0.529               & \textbf{0.640}      \\
Ethnicity (binary, AUC)             & 0.652               & 0.678               & ---                 & \textbf{0.860}      & 0.664               & 0.715               \\
APOE e4 carrier (binary, AUC)       & 0.637               & 0.639               & ---                 & 0.395               & 0.484               & \textbf{0.655}      \\
APOE genotype (3-way, AUC)          & 0.583               & 0.580               & ---                 & 0.480               & 0.491               & \textbf{0.600}      \\
Race (binary, AUC)                  & 0.904               & 0.906               & ---                 & 0.837               & 0.756               & \textbf{0.913}      \\
Scanner vendor (binary, AUC)        & \textbf{0.993}      & 0.993               & ---                 & ---                 & ---                 & 0.983               \\
\midrule
IDH1 mutation (AUC)                  & 0.916               & 0.905               & \textbf{0.937}      & 0.905               & 0.931               & 0.928               \\
MGMT methylation (AUC)              & 0.641               & 0.622               & 0.647               & \textbf{0.656}      & 0.643               & 0.653               \\
Tumor grade (3-way, AUC)           & 0.937               & 0.921               & \textbf{0.959}      & 0.936               & 0.949               & 0.951               \\
CDR staging (5-way, AUC)            & 0.717               & 0.717               & ---                 & ---                 & ---                 & \textbf{0.761}      \\
\midrule
Brain age ($R^{2}$)                 & 0.942               & \textbf{0.954}      & 0.692               & 0.891               & 0.699               & 0.952               \\
MMSE ($R^{2}$)                      & \textbf{0.427}      & 0.416               & ---                 & $-0.074$            & ---                 & 0.374               \\
CDR continuous ($R^{2}$)            & \textbf{0.316}      & \textbf{0.316}      & ---                 & ---                 & ---                 & 0.300               \\
MoCA ($R^{2}$)                      & 0.234               & 0.218               & ---                 & $-0.217$            & $-0.224$            & \textbf{0.285}      \\
Education years ($R^{2}$)           & 0.118               & 0.130               & ---                 & $-0.303$            & $-0.232$            & \textbf{0.455}      \\
GDS depression ($R^{2}$)            & $-0.103$            & $-0.098$            & ---                 & $-0.402$            & $-0.269$            & \textbf{$-0.038$}   \\
UPDRS-III ($R^{2}$)                 & $-0.166$            & $-0.153$            & ---                 & $-0.275$            & $-0.075$            & \textbf{$-0.017$}   \\
SCOPA autonomic ($R^{2}$)           & $-0.022$            & $-0.012$            & ---                 & $-0.392$            & $-0.119$            & \textbf{0.177}      \\
\bottomrule
\end{tabular}%
}
\end{table}

\section{Head-to-head competitor probing — full 23-task breakdown}
\label{app:competitors-full}

Complement to Table~\ref{tab:positioning}: all 23 probed tasks, four frozen encoders on identical splits and probe code (\Ncorpus{}-volume corpus, 5-fold stratified subject-grouped CV). Each bar shows the best-performing modality slice for that encoder on that task, so each encoder is reported at its strongest configuration. Table~\ref{tab:positioning} fixes the modality slice across encoders on 8 representative tasks for a directly comparable head-to-head.

\begin{figure}[H]
  \centering
  \includegraphics[width=\linewidth]{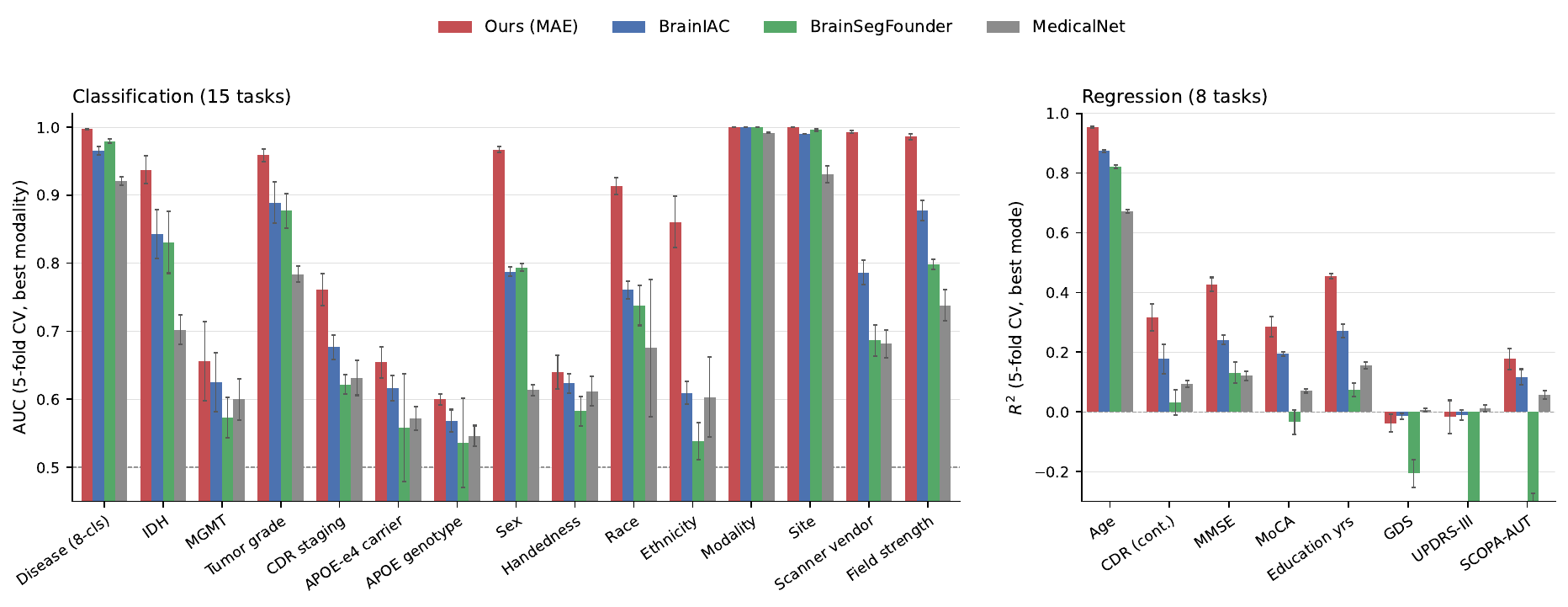}
  \caption{Frozen-feature linear probing across all 23 tasks, best modality per encoder. Left: 15 classification tasks, AUC ($\uparrow$). Right: 8 regression tasks, $R^{2}$ ($\uparrow$). Ours outperforms or matches every competitor on 21 of 23 tasks; the two non-wins (GDS depression, UPDRS-III) are near-floor regression for every encoder.}
  \label{fig:competitors-full}
\end{figure}

\section{CFG scale sensitivity}
\label{app:cfg-sweep}

We sweep CFG scale $s \in \{1.5,\,2.0,\,3.0\}$ at $n{=}32$ samples per arm (Table~\ref{tab:cfg-sweep}). CFG $s{=}1.5$ and $s{=}2.0$ are within noise on every controllable condition; at $s{=}3.0$ the continuous age axis degrades (Pearson $r$ drops $0.94 \to 0.82$) as strong guidance over-extrapolates away from the conditional manifold. IDH1 stays near chance throughout the sweep (mean agreement $0.51$--$0.52$), reflecting CFG's difficulty steering toward the rare mutant class on the small T1c tumor subset rather than a CFG-scale effect. We use $s{=}2.0$ throughout the paper as a point on the near-optimal plateau.

\begin{table}[H]
\centering
\small
\caption{Probe recovery of conditional attributes under different CFG scales. Columns: probe agreement with requested class (disease HC-vs-AD, sex F-vs-M, IDH1 wt-vs-mut) and Pearson $r$ between requested and predicted age on the 30\,/\,50\,/\,70 sweep, with the per-requested-age predicted means.}
\label{tab:cfg-sweep}
\begin{tabular}{lccccc}
\toprule
CFG $s$ & Disease & Sex  & Age $r$ & Age pred (30\,/\,50\,/\,70)\,y   & IDH1  \\
\midrule
1.5     & 1.00    & 0.92 & 0.95    & 30.9\,/\,49.4\,/\,72.4           & 0.52 \\
\textbf{2.0}
        & 0.98    & 0.92 & 0.94    & 35.4\,/\,54.8\,/\,75.7           & 0.52 \\
3.0     & 1.00    & 0.97 & 0.82    & 47.9\,/\,58.3\,/\,78.5           & 0.51 \\
\bottomrule
\end{tabular}
\end{table}

\section{Generation fidelity --- full decomposition}
\label{app:gen}

\paragraph{Protocol.} 3D-FID is computed via Inception-V3 on 2D slices. For each volume we extract $5$ central slices per anatomical view ($\times 3$ views) $=$ $15$ slices/volume. Real reference sets are drawn from the \Ncorpus{}-volume corpus under the stated filter; generated samples are drawn from the conditional DiT under CFG $s{=}2.0$, $50$-step Euler ODE. For each controllability arm we report three FIDs: \texttt{gen-vs-real} (generated vs real volumes), \texttt{gen-vs-recons} (generated vs the same real volumes passed through the tokenizer and reconstructed), and \texttt{recons-vs-real} (reconstructions vs real, i.e.\ the tokenizer floor).

\begin{table}[H]
\centering
\small
\caption{FID decomposition across controllability arms. \texttt{gen-vs-recons} measures how well the generator matches the tokenizer's conditional latent distribution; \texttt{recons-vs-real} is the tokenizer floor. Lower is better.}
\label{tab:fid-full}
\begin{tabular}{lrrr}
\toprule
Arm                        & gen-vs-real  & gen-vs-recons & recons-vs-real \\
\midrule
Disease HC (T1)            & 137.0        &  31.8         & 112.6          \\
Disease AD (T1)            & 135.6        &  25.5         & 124.7          \\
Sex F (T1)                 & 114.8        &  37.4         & 106.6          \\
Sex M (T1)                 & 117.7        &  54.6         &  93.4          \\
Modality T1                & 118.7        &  53.3         &  94.0          \\
Modality T2                & 100.6        &  64.6         &  79.8          \\
Modality FLAIR             & 111.7        &  55.6         &  89.2          \\
Modality T1c               & 133.1        &  \textbf{15.8} & 121.0         \\
Age $\approx$30 (T1)       & 123.8        &  27.7         & 118.3          \\
Age $\approx$50 (T1)       & 121.5        &  32.6         & 109.2          \\
Age $\approx$70 (T1)       & 122.5        &  29.1         & 111.2          \\
IDH1 wildtype (T1c, GBM)    & 133.8        &  \textbf{15.8} & 121.3         \\
IDH1 mutant (T1c, GBM)      & 131.2        &  \textbf{22.2} & 123.2         \\
\midrule
T1 pooled                  & 118.6        &  53.4         &  94.0          \\
\emph{Pooled across all arms} & \textbf{107.3} & \textbf{34.4} & \textbf{77.4} \\
\bottomrule
\end{tabular}
\end{table}

\paragraph{Memorization audit.} Nearest-neighbor L$_{2}$ distances computed in the $1200{\times}32$ token latent ($38{,}400$-dim flattened), real pool $n{=}33{,}639$, synthetic $n{=}1088$. Real-to-real NN: mean $37.2$, median $33.4$, $5^{\text{th}}$ percentile $14.4$, min $2.95$, max $116.3$. Synthetic-to-real NN: mean $57.7$, median $60.4$, min $23.6$, max $75.0$. Ratio of means $=1.55$; zero synthetic samples fall below the $5^{\text{th}}$-percentile real-to-real threshold. Verdict: no memorization.

\begin{figure}[H]
  \centering
  \includegraphics[width=0.8\linewidth]{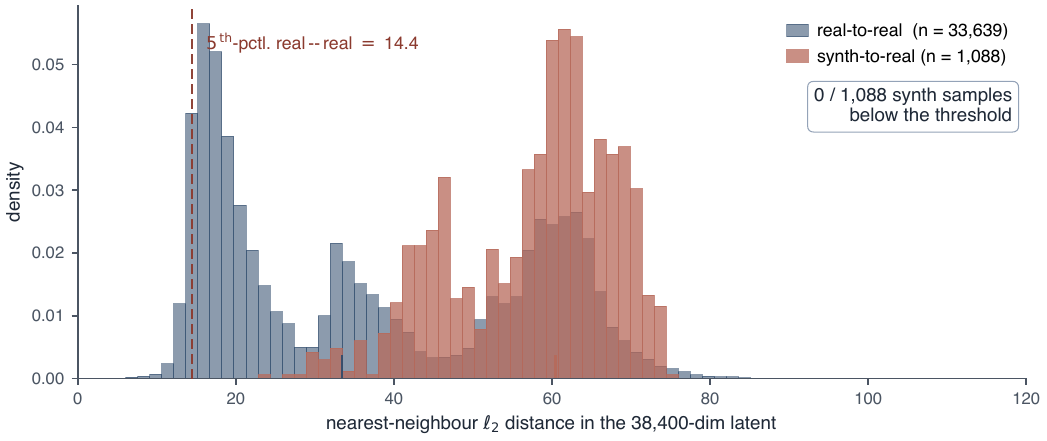}
  \caption{Nearest-neighbor L$_{2}$ distributions in the $38{,}400$-dim latent space. Real-to-real (slate) versus generated-to-real (coral). The synthetic distribution sits entirely to the right of the real distribution's lower tail; zero of $1088$ generated samples falls below the $5^{\text{th}}$-percentile real-to-real threshold (dashed line, $14.4$).}
  \label{fig:memhist}
\end{figure}

\section{Longitudinal sweep visualization}
\label{app:longit-sweep}

\begin{figure}[H]
  \centering
  \includegraphics[width=0.95\linewidth]{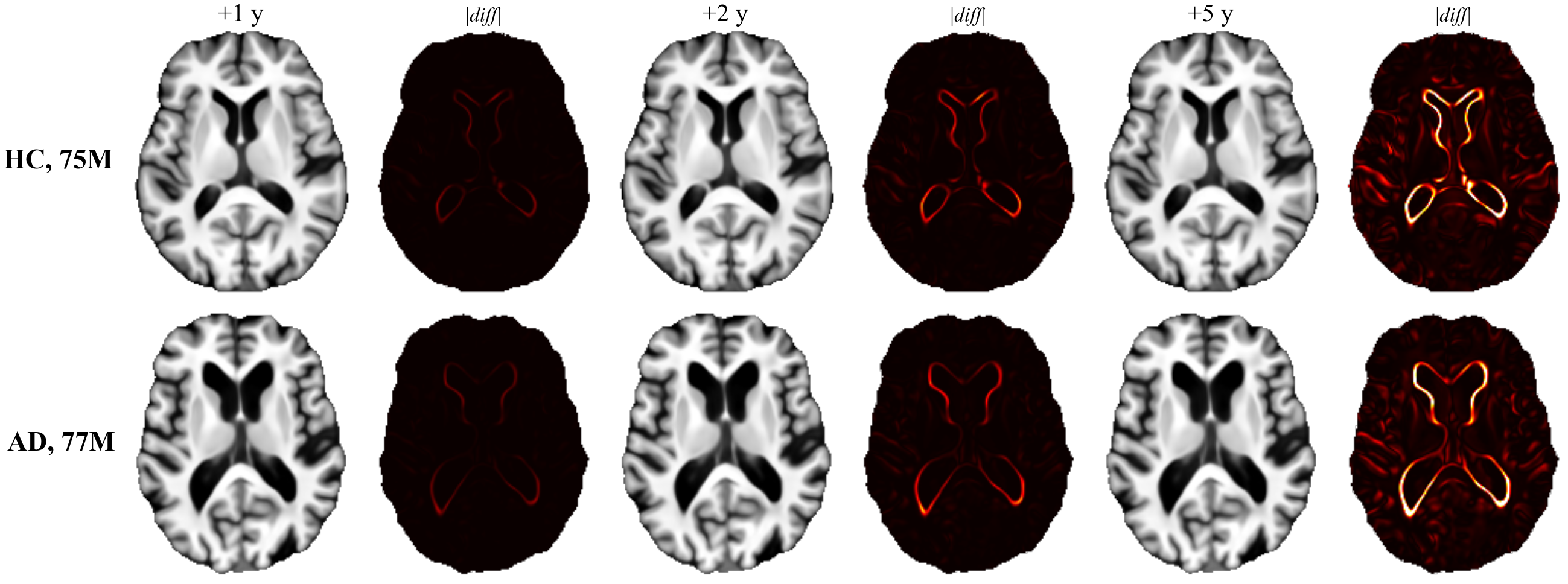}
  \caption{Longitudinal forecasting at fixed baseline. Two held-out ADNI cases (HC 75-yr-old male, AD 77-yr-old male) bridge-sampled at requested $\Delta t \in \{1, 2, 5\}$\,y. Each pair: forecasted axial slice (left) and the absolute voxel difference $|\,\text{sample}(\Delta t) - \text{sample}(0)\,|$ (right, hot colormap, $\gamma{=}1.5$); subtracting the round-trip noise floor isolates $\Delta t$-driven structural change. Difference maps grow with requested $\Delta t$ and concentrate along cortex and ventricles, the expected loci of age-related change. T1.}
  \label{fig:longit}
\end{figure}

Complement to \S\ref{sec:exp-longitudinal}: per-$\Delta t$ self-consistency of the bridge sampler. The same two held-out ADNI cases shown in Figure~\ref{fig:realvsforecast} are bridge-sampled at requested $\Delta t \in \{1, 2, 5\}$\,y, and each forecast is compared to the $\Delta t{=}0$ sample to isolate $\Delta t$-driven structural change (subtracting the round-trip sampler noise floor). Figure~\ref{fig:longit} shows that the difference maps grow monotonically with requested $\Delta t$ and concentrate along the same cortex and ventricle loci that the real-vs-forecast comparison in Figure~\ref{fig:realvsforecast} highlights.

\end{document}